\documentclass[10pt,twocolumn,letterpaper]{article}

\usepackage{wacv}              

\usepackage{amsmath,amsfonts}
\usepackage{array}
\usepackage{textcomp}
\usepackage{stfloats}
\usepackage{url}
\usepackage{verbatim}
\usepackage{graphicx}
\usepackage{cite}
\usepackage[pagebackref,breaklinks,colorlinks]{hyperref}
\usepackage{overpic}
\usepackage{wrapfig}
\usepackage{cuted}
\usepackage{capt-of}
\usepackage{colortbl}
\usepackage{xcolor}
\usepackage{array}
\usepackage{graphicx}
\usepackage{lipsum}
\usepackage{arydshln}
\usepackage{xspace}
\usepackage{multirow}
\usepackage{multicol}
\usepackage{booktabs} 
\usepackage{tabu}
\usepackage[accsupp]{axessibility}
\usepackage{makecell}
\usepackage{mwe}
\usepackage{bm}
\usepackage[normalem]{ulem} 
\usepackage{enumitem}
\setlist[itemize]{noitemsep, nolistsep}
\usepackage{soul}

\newcommand{\del}[1]{}

\def\wacvPaperID{459} 
\def\confName{WACV}
\def\confYear{2025}

\definecolor{colorTrd}{rgb}{0.95,0.95,0.65}
\definecolor{colorSnd}{rgb}{1, 0.85, 0.7}
\definecolor{colorFst}{rgb}{1, 0.7, 0.7}
\definecolor{green}{rgb}{0.45, 0.62, 0.31}
\definecolor{blue}{rgb}{0.42, 0.60, 0.82}
\definecolor{yellow}{rgb}{0.80, 0.64, 0.22}

\begin{document}

\title{TriaGS: Differentiable Triangulation-Guided Geometric Consistency for 3D Gaussian Splatting}

\author{Quan Tran\\
VinUniversity\\
Ha Noi, Vietnam \\
{\tt\small quan.th@vinuni.edu.vn}
\and
Tuan Dang\\
University of Arkansas \\
Fayetteville, Arkansas, USA\\
{\tt\small tuand@uark.edu}
}

\twocolumn[{%
        \renewcommand\twocolumn[1][]{#1}
	\maketitle
        \vspace{0pt}
	\begin{center}
		\centering
            \begin{overpic}[width=\linewidth]{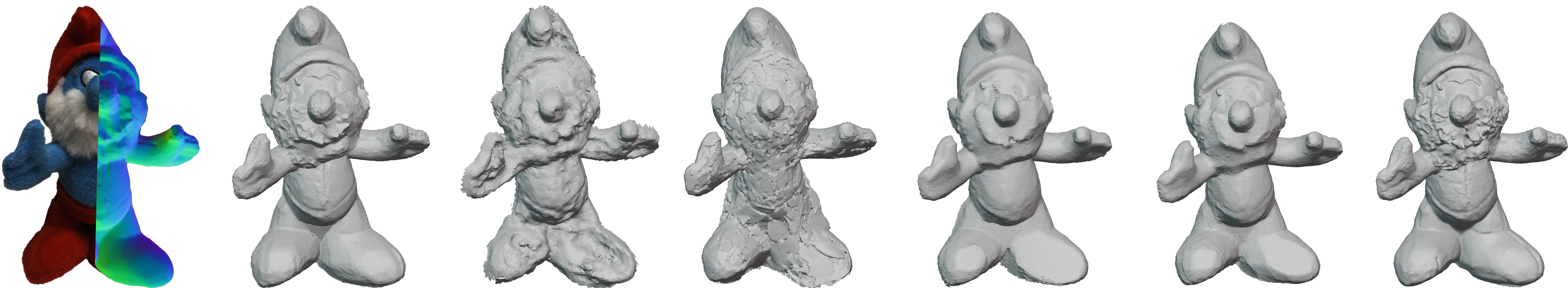}
                \put(0.5, -2.5){\small (a) Rendering}
                \put(17.5, -2.5){\small (b) Ours}
                \put(31.5, -2.5){\small (c) 3DGS}
                \put(45.5,-2.5){\small (e) Sugar}
                \put(60.5,-2.5){\small (f) 2DGS}
                \put(75,-2.5){\small (g) RaDe-GS}
                \put(90,-2.5){\small (h) PGSR}
          \end{overpic}
          \vspace{0pt}
	   \captionof{figure}{\textbf{TriaGS achieves superior surface reconstruction quality}. Our method (b) successfully creates a smooth and detailed surface, addressing a trade-off in reconstructions from GS. TriaGS eliminates the "lumpy" surface artifacts seen in 3DGS (c) and SuGaR (e) while preserving finer details than over-smoothing methods like 2DGS (f), RaDe-GS (g), and PGSR (h). Our quality is achieved through global geometric consistency constraint, which is based on differentiable triangulation. Sub-figure (a) shows the rendered color and normal map.}
	\label{fig:teaser}
	\end{center}
}]

\begin{abstract}
3D Gaussian Splatting is crucial for real-time novel view synthesis due to its efficiency and ability to render photorealistic images. However, building a 3D Gaussian is guided solely by photometric loss, which can result in inconsistencies in reconstruction. This under-constrained process often results in "floater" artifacts and unstructured geometry, preventing the extraction of high-fidelity surfaces. To address this issue, our paper introduces a novel method that improves reconstruction by enforcing global geometry consistency through constrained multi-view triangulation. Our approach aims to achieve a consensus on 3D representation in the physical world by utilizing various estimated views. We optimize this process by penalizing the deviation of a rendered 3D point from a robust consensus point, which is re-triangulated from a bundle of neighboring views in a self-supervised fashion. We demonstrate the effectiveness of our method across multiple datasets, achieving state-of-the-art results. On the DTU dataset, our method attains a mean Chamfer Distance of $0.50$ mm, outperforming comparable explicit methods. We will make our code open-source to facilitate community validation and ensure reproducibility.
\end{abstract}

\section{Introduction}
3D Gaussian Splatting (3DGS) has transformed real-time novel view synthesis with its exceptional rendering quality and efficiency \cite{3dgs}. By representing scenes as a collection of explicit Gaussian primitives, 3DGS replaces the costly ray-marching used in Neural Radiance Fields (NeRFs) \cite{nerf} with a differentiable rasterization pipeline. Despite its rendering success, extracting a high-fidelity and geometrically accurate surface mesh remains a significant challenge. This difficulty stems from the representation itself, as the optimization of 3DGS is guided solely by photometric loss, which prioritizes novel view synthesis over geometric accuracy. Consequently, semi-transparent Gaussian "floaters" in space can be composited to produce the target image without adhering to a single coherent surface \cite{2024sugar, 20243dgsr}. This under-constrained process presents a significant challenge in extracting a high-fidelity surface.

Recent works have introduced geometric regularization-based methods to tackle this issue. Notably, some approaches encourage the Gaussians to be planar \cite{neusg}, and enforce consistency between pairs of camera views \cite{gaussurf, gaussianpro, pgsr}. However, such pairwise supervision is strictly local, making it prone to error accumulation and global geometric drift (Figure \ref{fig:pairwise}). This limitation hinders downstream applications, such as physics-based simulations and robotics, that require precise geometry. In contrast, multi-view triangulation aggregates constraints across several views simultaneously, enabling outlier rejection and offering a principled mechanism for disambiguation.

\begin{figure*}[htbp]
  \centering
  \includegraphics[width=\textwidth]{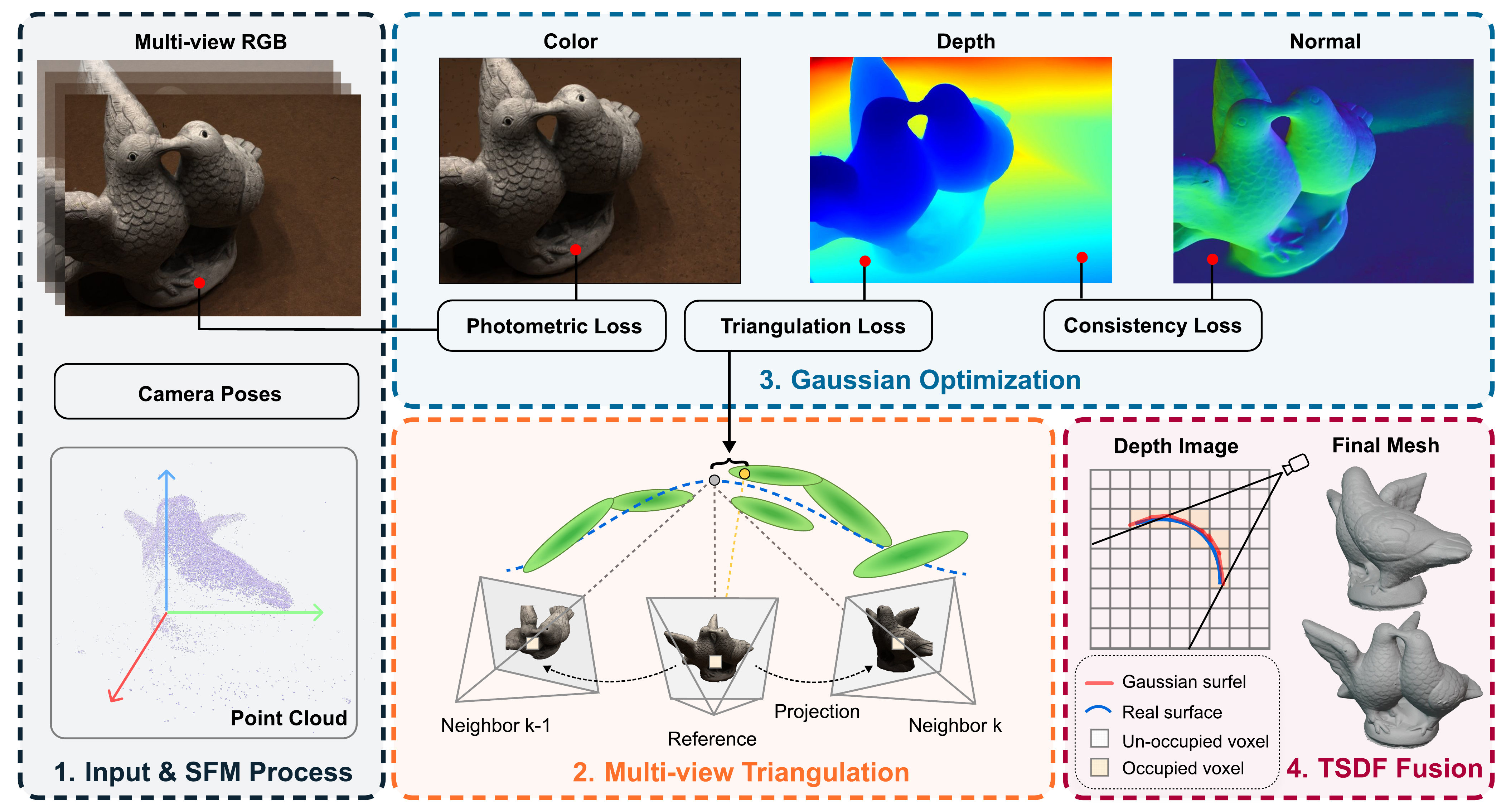}
  \caption{\textbf{An overview of the TriaGS reconstruction pipeline}. Our method begins with a standard SfM initialization (1). Our main contribution is a differentiable multi-view triangulation module (2) that enforces global geometric consistency. For a rendered point, we re-triangulate a robust consensus point from neighboring views. The residual between these points forms our novel Triangulation Loss, which guides the Gaussian Optimization (3). Finally, we extract the final high-fidelity mesh via TSDF fusion of rendered depth maps (4).}
  \label{fig:pipeline}
\end{figure*}

To address this, we argue that robust global geometry is better achieved through collective multi-view consensus. We introduce TriaGS, a novel framework that implements this principle through classical multi-view triangulation in a fully self-supervised manner. Instead of comparing two views, our method assesses a rendered 3D point by re-triangulating its position from a bundle of neighboring views. By minimizing the distance between the initial estimate and this consensus point using a differentiable loss, our method guides the optimization toward a globally coherent surface. Our main contributions are:
\begin{itemize}[noitemsep]
    \item We introduce Triangulation-Guided Geometric Consistency, a novel, self-supervised loss function that utilizes classical multi-view triangulation to enforce global geometric consistency, moving beyond the limitations of local pairwise consistency checks.
    \item We propose a fully differentiable framework that directly integrates this triangulation-based consensus mechanism into the 3DGS optimization pipeline, providing a robust geometric signal to guide the primitives toward an accurate surface representation.
    \item Our method achieves state-of-the-art surface reconstruction quality on challenging benchmarks. TriaGS demonstrates superior surface reconstruction quality compared to baselines (Figure \ref{fig:teaser}), where finer geometric details are preserved on complex scenes.
\end{itemize}
\section{Related Work}

We position our contribution within two primary research trends: geometric regularization and multi-view geometry for supervision. Recent efforts to regularize 3DGS geometry fall into two main categories. The first category directly enforces local geometric priors on Gaussian primitives, starting with planar constraints \cite{neusg, gaussian_surfels, 2dgs} and later adding consistency checks between pairs of camera views \cite{pgsr, gaussurf, gaussianpro, vcrgaus}. The second group involves hybrid approaches that jointly optimize a continuous implicit field, a Signed Distance Function (SDF), alongside the explicit Gaussians \cite{gs-pull, 2024gsdf, 20243dgsr, geoneus}. While these methods have advanced the state-of-the-art, they share a common limitation: they either rely on local supervision, which can be susceptible to error accumulation and global geometric drift, or they introduce the computational overhead and complexity of a secondary neural network. In contrast, our method remains within the 3DGS framework and enforces a global form of consistency. To achieve this, our approach is inspired by the foundational principle of using multi-view triangulation \cite{hartley2003multiple} for geometric validation. We draw from the concept of a differentiable triangulation loss, which has been used in tasks like pose estimation by minimizing the residual between an estimated 3D point and an optimal point triangulated from multiple 2D keypoints \cite{triangulation_residual_loss}. However, these methods require pre-matched 2D features or external modules, such as a pre-trained MVS network \cite{gs2mesh, depth_regularized, vcrgaus}. Our key distinction is that we adapt this multi-view consensus principle to be fully self-supervised. We enforce global geometric coherence by minimizing the discrepancy between a rendered 3D point and its optimal re-triangulation from multiple views. This directly addresses the geometric drift inherent in local pairwise checks and achieves global consistency without needing external supervisory signals.
\section{System Overview}

Our method introduces a novel geometric regularizer into the 3DGS pipeline to enforce global multi-view consistency. The pipeline, illustrated in Figure \ref{fig:pipeline}, transforms multi-view images with corresponding camera poses into a high-fidelity surface mesh. The pipeline begins by initializing a cloud of 3D Gaussian primitives initialized using sparse points from Structure-from-Motion (SfM) \cite{structure-from-motion}. The primary challenge we address is that standard 3DGS rendering is optimized for photorealism rather than geometric accuracy, resulting in inconsistent depth maps across different views. We leverage the physically-grounded formulation from RaDe-GS \cite{radegs} to render consistent depth and normal maps, which provide a reliable 3D point for each pixel. Using this reliable point as an anchor, we introduce our key contribution: a self-supervised check for global geometric agreement. For a point on the surface rendered from a reference view, we evaluate its global integrity by re-projecting it into a bundle of neighboring camera views. This creates a classical triangulation problem. We solve this problem using a differentiable Singular Value Decomposition (SVD) to obtain a new geometrically robust "consensus" point. Our novel Triangulation-Guided Geometric Consistency (TGGC) loss then penalizes the distance between the initial rendered estimate and this multi-view consensus point. This geometrically grounded loss guides the entire Gaussian cloud toward a single coherent surface. Finally, after training converges, we extract the final high-quality mesh by fusing clean depth maps rendered from all training views into a Truncated Signed Distance Function (TSDF).

\section{Methodology}

\subsection{Rasterizing Consistent Depth and Normals}
\label{section:rasterizer_depth}
3DGS \cite{3dgs} models a scene using a collection of 3D Gaussian primitives. Each Gaussian, $G_i$, is defined by a mean (center) $\boldsymbol{\mu} \in \mathbb{R}^3$, a covariance matrix $\boldsymbol{\Sigma} = R S S^T R^T$, an opacity $\alpha$, and Spherical Harmonics (SH) coefficients. The final pixel color is synthesized by alpha-blending the contributions of all overlapping, depth-sorted Gaussians \cite{zwicker2001ewa}. Similarly, the standard rendered depth is the alpha-blended depth of the Gaussian centers:
\begin{equation}
\label{eq:render_depth}
D(\mathbf{u}) = \sum_{i=1}^N T_i \alpha_i' z_i, \quad \text{where } T_i = \prod_{j=1}^{i-1}(1-\alpha_j').
\end{equation}

\begin{figure}[t]
    \centering
    \captionsetup{font=footnotesize}
    \includegraphics[width=\linewidth]{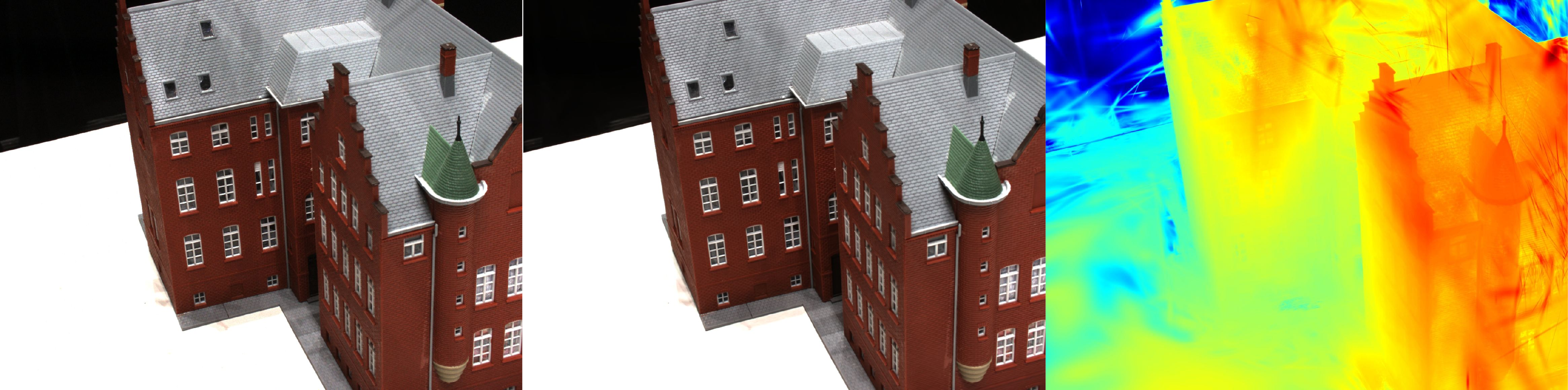}
    \caption{
        \textbf{Geometric inconsistency in standard 3DGS depth rendering}. While matching the ground truth image (left) photometrically (middle), the standard depth map (right) is noisy and fails to capture the scene's geometry, making it an unreliable foundation for reconstruction.
    }
    \label{fig:floater_artifact}
\end{figure}

However, this formulation is geometrically flawed. Blending the depths of Gaussian centers ignores the shape and orientation of the Gaussian primitives. For example, a ray intersecting a slanted, disk-like Gaussian will produce a depth value that does not correspond to any actual surface point it represents. This incongruity is a major source of noisy depth maps and "floater" artifacts (illustrated in Figure \ref{fig:floater_artifact}), making this depth map rendering disconnected from the true surface geometry and creating an unreliable anchor for high-fidelity geometric supervision, which would undermine multi-view triangulation. In our work, we require a accurate 3D point for each pixel to form a triangulation. To overcome this, we build upon the rasterizable depth formulation introduced in RaDe-GS \cite{radegs}. This method employs a physically grounded approach to determine the precise intersection of a viewing ray with a 3D Gaussian ellipsoid, resulting in a geometrically consistent depth and normal map. Particularly, the intersection point between a viewing ray and a Gaussian is defined as the location of maximum density along that ray. For a ray parameterized by distance $t$ from the camera origin $\mathbf{o}$ as $\mathbf{x}(t) = \mathbf{o} + t\mathbf{v}$, its maximum occurs at the intersection distance $t^*$:
\begin{equation}
\label{eq:t_star_perspective}
t^* = \frac{\mathbf{v}^T \boldsymbol{\Sigma}^{-1} (\mathbf{x}_c - \mathbf{o})}{\mathbf{v}^T \boldsymbol{\Sigma}^{-1} \mathbf{v}}
\end{equation}

\noindent where $\mathbf{x}_c$ and $\boldsymbol{\Sigma}$ denote the Gaussian's center and covariance. This equation presents a challenge: computing the direct $t^*$ value for every pixel is too complex for real-time applications because the viewing direction $\mathbf{v}$ changes per pixel. RaDe-GS resolves this by leveraging the local affine approximation already present in the 3DGS rasterizer. This transforms the problem from the perspective "camera space" into a local non-Cartesian "ray space," where all viewing rays are treated as parallel. The complex intersection formula (Eq. \ref{eq:t_star_perspective}) collapses into a simple dot product:
\begin{equation}
t^* = \mathbf{\hat{q}} \cdot (\mathbf{u}_c - \mathbf{u}_o), \quad \text{where} \quad \mathbf{\hat{q}} = \frac{\mathbf{v}'^T \boldsymbol{\Sigma}'^{-1}}{\mathbf{v}'^T \boldsymbol{\Sigma}'^{-1}\mathbf{v}'}.
\end{equation}

Here, $\mathbf{u}_c$ and $\mathbf{u}_o$ are the Gaussian center and pixel locations in ray space, respectively. Since $\mathbf{\hat{q}}$ can be pre-computed for each Gaussian, finding the intersection distance $t^*$ becomes fully rasterizable. From the intersection distance $t^*$, the actual depth $d$ is computed as its orthogonal projection onto the camera's principal axis, approximated using the angle of the Gaussian center $\theta_c$: $d = t^* \cos\theta_c$.

The normal vector, denoted as \( \mathbf{n}' \), is calculated from the parameters of the transformed Gaussian. Since the intersection points in ray space form a plane, we can easily determine its normal. The normal vector $n'$ is derived from the transformed Gaussian's parameters, then transformed back to the camera space using the Jacobian of the local affine transformation, $J$: $\mathbf{n} = J^T \mathbf{n}'$.

By rendering in this manner, we obtain a precise 3D point that lies on a locally consistent planar surface. Rade-GS provides a robust, geometrically sound anchor for our primary contribution: a multi-view triangulation consistency check, defined in the next section.

\subsection{Differentiable Multi-View Triangulation}

\begin{figure}[t]
    \includegraphics[width=\linewidth]{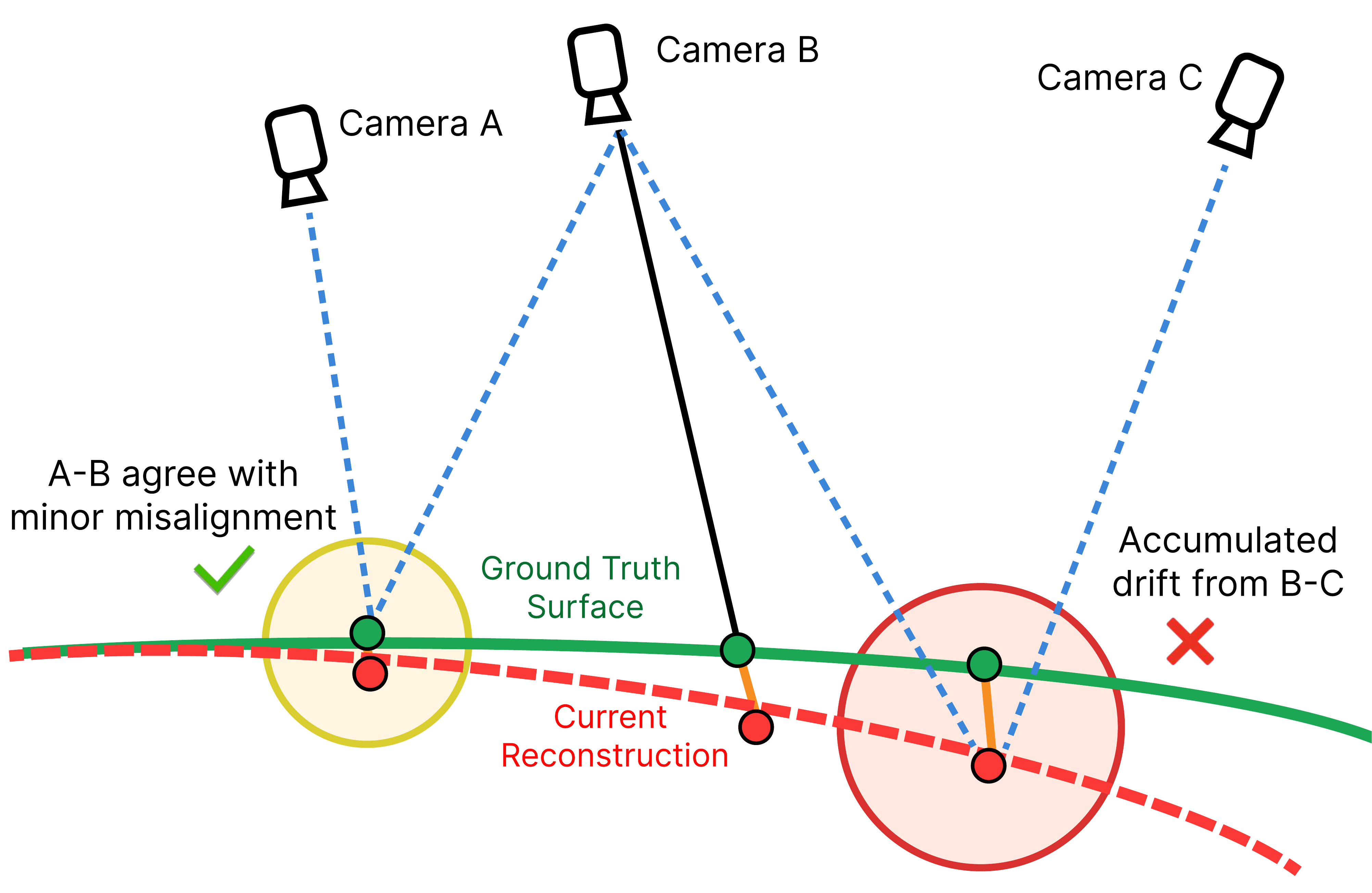}
    \captionsetup{font=footnotesize}
    \centering
        \caption{\textbf{Limitation of Pairwise Consistency} Local pairwise checks can accumulate errors, leading to global geometric drift. A check between views A and B may show only a "minor misalignment" and be considered consistent. However, this error compounds in subsequent checks (e.g., B to C), resulting in a significant "accumulated drift". This causes reconstructed surface (dashed red line) to deviate from the ground truth. Our method avoids this failure by enforcing a simultaneous multi-view consensus.
    }
    \label{fig:pairwise}
    \vspace{-15pt}
\end{figure}

Recent methods for geometric regularization in 3DGS \cite{pgsr,gaussianpro,gaussurf} focus on enforcing multi-view consistency through pairwise comparisons between camera views. These losses often minimize geometric re-projection errors or photometric similarities between adjacent views. However, they can be susceptible to error accumulation and may struggle to resolve geometric ambiguities when trying to achieve consensus across a broader set of camera viewpoints. The primary issue with iterative optimization is that the rendered geometry at any given training step is merely an estimate, not the actual ground truth. As illustrated in Figure \ref{fig:pairwise}, this local supervision with incomplete geometry can lead to minor misalignments that accumulate through a series of pairwise comparisons, eventually resulting in significant global geometric drift. Such drift prevents the reconstruction of a coherent surface, which is crucial for applications beyond novel view synthesis. We address this limitation by introducing Triangulation-Guided Geometric Consistency (TGGC) (see section \ref{section:tggc}), a novel self-supervised regularization loss inspired by classical multi-view triangulation principles. Instead of verifying consistency between two views, our method assesses the geometric integrity of a rendered 3D point by measuring its self-consistency against an optimal re-triangulated point from a bundle of its viewing rays. Our loss provides a more robust geometric signal guiding optimization toward a global surface.

We first establish a consistent mathematical representation for all cameras in the scene. The pose of any camera view $i$ is obtained from an initial Structure-from-Motion (SfM) process \cite{structure-from-motion}. Each pose is described by a shared intrinsic matrix $K$ and a rigid body transformation $\mathbf{M}_{i}$ that maps coordinates from the world frame to the camera's local frame. For a specific pixel $p_r = [u_r, v_r]^T$ in a reference camera view $r$, we first render its depth $d_r$ and normal $n_r$ using the rasterizable formulation described in section \ref{section:rasterizer_depth}. In conjunction with the pre-calibrated camera poses, these rendered parameters are then used to compute a current 3D surface point, $X_r$, by intersecting the viewing ray corresponding to pixel $p_r$ with this plane. This point is a differentiable function of the optimizable Gaussian attributes. While this point is valid geometrically for the current rendering of the reference view, its consistency with other viewpoints is not guaranteed. If $X_r$ accurately represents the actual surface location, it must also be consistent with observations from different perspectives.

We formalize a differentiable multi-view triangulation mechanism to enforce global consistency.  We first identify a set of $k$ neighboring views surrounding the reference view $ r$. We then differentially project the current 3D point $X_r$ onto the image planes of all $k+1$ views using its corresponding $3 \times 4$ projection matrix $P_i$. This step generates a set of $k+1$ corresponding 2D pixel observations, which we denote as $\{p'_i = [u'_i, v'_i]^T\}$:
\begin{equation}
\label{eq:projection_pi_prime}
p'_i = s_i \begin{bmatrix} u_i' & v_i' & 1 \end{bmatrix}^T = P_i \begin{bmatrix} x_r & y_r & z_r & 1 \end{bmatrix}^T
\end{equation}

\noindent where $s_i$ is the projective depth, and $P_{ij}^T$ is the $j$-th row of the projection matrix $P_i$. We highlight that these 2D points $\{p'_i\}$ are obtained through the differentiable projection of $X_r$ onto the image planes of the selected views. They do not result from feature matching or photometric alignment between the input images. Our approach ensures that the entire process remains end-to-end differentiable with respect to the Gaussian attributes. If we were to deviate from this approach, it would introduce a non-differentiable operation, disrupt the gradient flow, and render end-to-end optimization impossible. Therefore, our method relies exclusively on a geometric self-consistency check, eliminating the need for a separate model.

The differentiable projection of $X_r$ onto the image planes of all $k+1$ views results in a set of corresponding 2D points and projection matrices, $\{(p'_i, P_i)\}_{i=1}^{k+1}$. This collection now forms a well-posed input for classical multi-view triangulation. We follow the Direct Linear Transform method \cite{abdel2015direct}, where each observation pair $(p'_i, P_i)$, with $p'_i = [u'_i, v'_i]^T$, imposes two linear constraints on the homogeneous coordinates of the unknown 3D point $X$: 
\begin{equation}
\label{eq:linear_system_AX0}
\begin{split}
(u_i' P_{i3}^T - P_{i1}^T) X = 0 \text{ and } (v_i' P_{i3}^T - P_{i2}^T) X = 0
\end{split}
\end{equation}

\noindent where $P_{ij}^T$ denotes the $j$-th row of the projection matrix $P_i$. Stacking these constraints for all $k+1$ views forms an overdetermined linear system $AX = \mathbf{0}$, where the $2(k+1) \times 4$ matrix $A$ encapsulates the multi-view geometric constraints derived from the current Gaussian state. Assuming that all projection matrices are perfectly estimated, each row of $A$, multiplied by $X$, must equal zero if $X$ perfectly lies on the corresponding ray. In an ideal, noise-free scenario where the $X_r$ is perfectly accurate to represent the true 3D surface point, all $k+1$ viewing rays would intersect precisely at $X_r$. From an algebraic perspective, this perfect intersection translates to the system of linear equations $AX=0$ having a non-trivial solution \cite{ROMEO202087}. This provides the algebraic foundation for our consistency check.

\subsection{The Triangulation-Guided Consistency Loss}
\label{section:tggc}

In practice, the $k+1$ viewing rays defined by the system $AX = \mathbf{0}$ do not intersect at a single point in 3D space for two fundamental reasons. First, the rendered 3D point, $X_r$, is merely the current optimization step's geometric estimate from a reference view, not the true surface location. Second, the input camera poses, derived from SfM, contain unavoidable estimation errors \cite{schoenberger2016sfm}. Consequently, when $X_r$ is projected into the $k+1$ views, the resulting rays are inconsistent, forming a system with no exact solution. Thus, we aim to seek the optimal "consensus point" $X^*$ that best satisfies all $k+1$ geometric constraints simultaneously by minimizing the algebraic error, represented by the residual $||AX||^2$. This classic problem can be solved using differentiable Singular Value Decomposition (SVD) of the matrix $A = U\Sigma V^T$. According to linear algebra results, the unit-norm vector minimizing $||AX||^2$ is the right singular vector of $A$ associated with the smallest singular value \cite{hartley2003multiple}. This vector is the last column of $V$, representing the re-triangulation point $X^*$, which is most geometrically consistent with the entire $k+1$ multi-view observations. \del{ Our implementation utilizes modern auto-differentiation libraries, PyTorch, to compute the full SVD operations.} We construct and solve for a batch of matrices $\{A_b\}_{b=1}^B$ in parallel at each iteration to amortize the computational cost of the SVD. The geometric residual $||X_r - X^*||$, is the gradient signal for our optimization. It measures the difference between $X_r$ and the re-triangulated point $X^*$. Our regularizer is therefore formulated to minimize this discrepancy. 

However, a standard $L_2$ loss is susceptible to outliers, due to its quadratic nature, particularly in the early stages of optimization where the renders are still noisy. This sensitivity can lead to explosive gradients in response to large errors, potentially destabilizing the training process. We instead employ the Geman-McClure loss function \cite{geman1986bayesian}, a robust error metric designed to mitigate the influence of outliers, which often arise from challenging scenarios like occlusions or views of textureless surfaces. The loss is defined as:
\begin{equation}
\label{eq:loss}
\mathcal{L}_{T} = \sum_{pixels} \frac{||X_r - X^*||_2^2}{||X_r - X^*||_2^2 + \sigma^2}
\end{equation}

\noindent where $\sigma$ is a scale parameter that controls its behavior. This loss behaves like a scaled $L_2$ norm for minor errors but saturates for large ones, preventing outliers from destabilizing the optimization. We anneal $\sigma$ throughout training via an exponential decay schedule to encourage high precision in the final stages. This entire process, from the SVD-based consensus point calculation to the loss $\mathcal{L}_{T}$, is fully differentiable. Gradients are propagated back to the Gaussian attributes as detailed in our supplementary material. Our approach introduces a differentiable variant of this classical geometric reasoning into the learning loop, enabling end-to-end optimization with a global geometric signal.

\begin{figure*}[h]
    \centering
    \begin{overpic}[width=\textwidth]{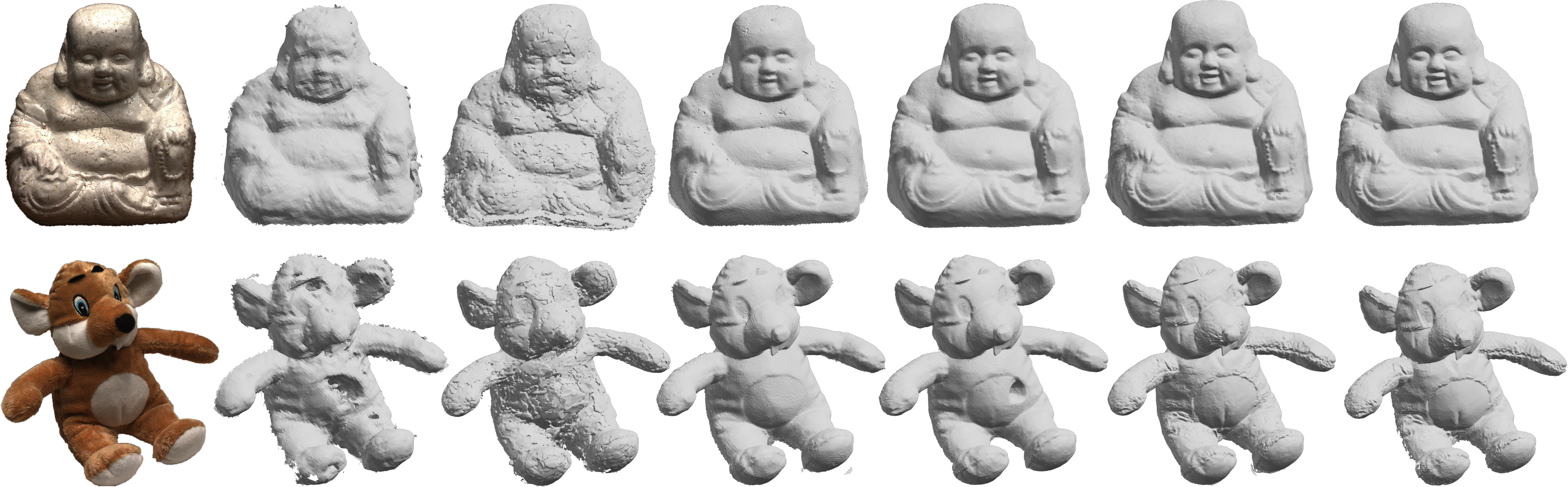}
        \put(3, -2.5){\small (a) Reference}
        \put(18, -2.5){\small (b) 3DGS}
        \put(33, -2.5){\small (c) Sugar}
        \put(46,-2.5){\small (e) 2DGS}
        \put(60,-2.5){\small (f) RaDe-GS}
        \put(75,-2.5){\small (g) PGSR}
        \put(90,-2.5){\small (h) Ours}
    \end{overpic}

    \captionsetup{font=footnotesize}
    \vspace{8pt}
    \caption{
        \textbf{Qualitative surface reconstruction on the DTU dataset}. Our method (h) yields visibly superior geometric fidelity, which produces smoother, more complete surfaces with finer details. In contrast, methods relying on local consistency, like PGSR (g), still exhibit "bumpy" artifacts.
    }
    \label{fig:dtu4_1}
\end{figure*}
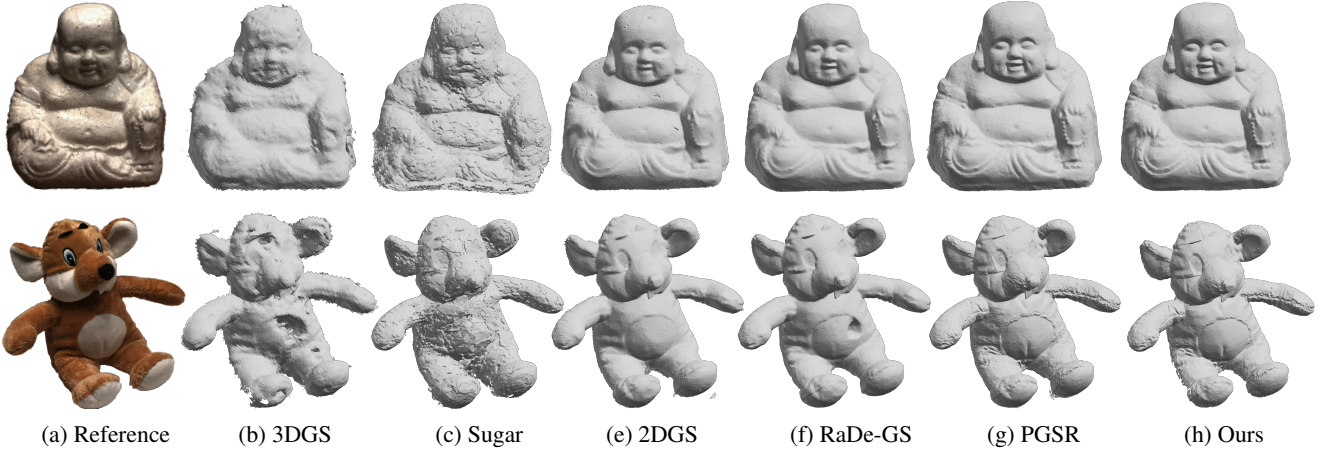

\begin{table*}[h]
\centering
\caption{Geometry quality comparison on the DTU dataset. We report an error metric (lower is better) compared with baselines. Mean scores and training times (m: minute, h: hour) are included at the bottom. Best results are highlighted as \colorbox{colorFst}{1st}, \colorbox{colorSnd}{2nd}, and \colorbox{colorTrd}{3rd}.}
\label{tab:dtu_geometry_comparison}
\resizebox{\textwidth}{!}{%
\begin{tabular}{l|ccccccccccccccc|c|r}
\toprule
\textbf{Method} & \textbf{24} & \textbf{37} & \textbf{40} & \textbf{55} & \textbf{63} & \textbf{65} & \textbf{69} & \textbf{83} & \textbf{97} & \textbf{105} & \textbf{106} & \textbf{110} & \textbf{114} & \textbf{118} & \textbf{122} & \textbf{Mean} & \textbf{Time} \\
\midrule
VolSDF & 1.14 & 1.26 & 0.81 & 0.49 & 1.25 & 0.70 & 0.72 & 1.29 & 1.18 & 0.70 & 0.66 & 1.08 & 0.42 & 0.61 & 0.55 & 0.86 & $>$12h \\
NeuS & 1.00 & 1.37 & 0.93 & 0.43 & 1.10 & 0.65 & 0.57 & 1.48 & 1.09 & 0.83 & 0.52 & 1.20 & 0.35 & 0.49 & 0.54 & 0.84 & $>$12h \\
Neuralangelo & \colorbox{colorTrd}{0.37} & \colorbox{colorTrd}{0.72} & 0.35 & \colorbox{colorTrd}{0.35} & 0.87 & \colorbox{colorFst}{0.54} & \colorbox{colorSnd}{0.53} & 1.29 & \colorbox{colorTrd}{0.97} & 0.73 & \colorbox{colorSnd}{0.47} & \colorbox{colorTrd}{0.74} & \colorbox{colorTrd}{0.32} & \colorbox{colorTrd}{0.41} & \colorbox{colorTrd}{0.43} & \colorbox{colorTrd}{0.61} & $>$12h \\
3DGS & 2.14 & 1.53 & 2.08 & 1.68 & 3.49 & 2.21 & 1.43 & 2.07 & 2.22 & 1.75 & 1.79 & 2.55 & 1.53 & 1.52 & 1.50 & 1.97 & 11.2m \\
SuGaR & 1.47 & 1.33 & 1.13 & 0.61 & 2.25 & 1.71 & 1.15 & 1.63 & 1.62 & 1.07 & 0.79 & 2.45 & 0.98 & 0.88 & 0.79 & 1.32 & 1h \\
2DGS & 0.48 & 0.91 & 0.39 & 0.39 & 1.01 & 0.83 & 0.81 & 1.36 & 1.27 & 0.76 & 0.70 & 1.40 & 0.40 & 0.76 & 0.52 & 0.80 & 0.32h \\
GOF & 0.50 & 0.82 & 0.37 & 0.37 & 1.12 & 0.74 & 0.73 & \colorbox{colorTrd}{1.18} & 1.29 & 0.68 & 0.77 & 0.90 & 0.42 & 0.66 & 0.49 & 0.74 & 2h \\
RadeGS & 0.46 & 0.73 & \colorbox{colorTrd}{0.33} & 0.38 & \colorbox{colorTrd}{0.79} & 0.75 & 0.76 & 1.19 & 1.22 & \colorbox{colorTrd}{0.62} & 0.70 & 0.78 & 0.36 & 0.68 & 0.47 & 0.68 & 10m \\
PGSR & \colorbox{colorFst}{0.34} & \colorbox{colorFst}{0.58} & \colorbox{colorSnd}{0.29} & \colorbox{colorFst}{0.29} & \colorbox{colorSnd}{0.78} & \colorbox{colorSnd}{0.58} & \colorbox{colorTrd}{0.54} & \colorbox{colorSnd}{1.01} & \colorbox{colorSnd}{0.73} & \colorbox{colorSnd}{0.51} & \colorbox{colorTrd}{0.49} & \colorbox{colorSnd}{0.69} & \colorbox{colorSnd}{0.31} & \colorbox{colorFst}{0.37} & \colorbox{colorSnd}{0.38} & \colorbox{colorSnd}{0.53} & 0.6h \\
\midrule 
\textbf{Ours} & \colorbox{colorSnd}{0.35} & \colorbox{colorFst}{0.58} & \colorbox{colorFst}{0.28} & \colorbox{colorSnd}{0.31} & \colorbox{colorFst}{0.72} & \colorbox{colorTrd}{0.62} & \colorbox{colorFst}{0.49} & \colorbox{colorFst}{0.97} & \colorbox{colorFst}{0.72} & \colorbox{colorFst}{0.48} & \colorbox{colorFst}{0.45} & \colorbox{colorFst}{0.46} & \colorbox{colorFst}{0.28} & \colorbox{colorSnd}{0.39} & \colorbox{colorFst}{0.34} & \colorbox{colorFst}{0.50} & \textbf{0.4h} \\
\bottomrule
\end{tabular}%
}
\end{table*}

\subsection{Optimization}
In addition to our Triangulation-Guided Loss \(L_{T}\), the model is unable to recover fine surface details if it is trained solely with the photometric supervision term. Our final objective function $\mathcal{L}$ is a weighted sum of these components: \del{, which is defined as}
\begin{equation}
\mathcal{L} = \mathcal{L}_{\text{photo}} + \lambda_{\text{t}}\mathcal{L}_{\text{T}} + \lambda_{\text{normal}}\mathcal{L}_{\text{normal}} + \lambda_{\text{photo-mv}}\mathcal{L}_{\text{photo-mv}}
\end{equation}

In this equation, \(\mathcal{L}_{\text{photo}}\) represents the primary photometric supervision term. We apply a local geometric regularizer, $\mathcal{L}_{\text{normal}}$, which contains a normal consistency term, following recent works \cite{gaussurf, 2dgs, radegs}. We also include an auxiliary multi-view photometric consistency loss, \(\mathcal{L}_{\text{photo-mv}}\), which contains a weighting term to handle potential occlusions. The \(\lambda\) coefficients are the corresponding loss weights, which are detailed in our supplementary material.
\section{Evaluations}

\subsection{Experiment Setup}

\textbf{Datasets and Metrics}: We evaluate surface reconstruction on standard benchmarks: the DTU dataset \cite{dtudataset}, comprising 15 object-centric scans, the large-scale Tanks and Temples (TNT) dataset \cite{tankstemplate}, and the NeRF-Synthetic dataset \cite{nerf}. For geometric quality, we report the Chamfer Distance (CD) in mm on DTU and the F1-Score on TNT, following official evaluation toolkits \cite{dtueval, tanksandtemples_eval}. We also assess rendering quality on the Mip-NeRF 360 \cite{mipnerf360} dataset using PSNR, SSIM, and LPIPS \cite{lpips}.

\textbf{Baselines}: We compare our method against a list of state-of-the-art approaches. Implicit baselines include NeuS \cite{neus} and the high-fidelity Neuralangelo \cite{neuralangelo}. Explicit Gaussian-based baselines include the 3DGS \cite{3dgs}, SuGaR \cite{2024sugar}, 2DGS \cite{2dgs}, GOF \cite{gof}, RaDe-GS \cite{radegs} and PGSR \cite{pgsr}.

\textbf{Implementation Details:} Our method uses a deterministic neighbor view selection policy. Our framework builds upon the official PyTorch implementation of 3D Gaussian Splatting \cite{3dgs}. For geometrically consistent rendering of depth and normals, we integrate the rasterizable formulation proposed by RaDe-GS \cite{radegs}. All experiments are conducted on a single NVIDIA RTX 4090 GPU. For all headline results, we use a fixed set of $k = 12$ neighbors per reference view. Our full training schedule is $30{,}000$ iterations, with the TGGC loss enabled after a $15{,}000$-iteration warm-up. 

\subsection{Geometric Reconstruction Quality}
\begin{figure*}[th]
    \centering
    \captionsetup{font=footnotesize}
    \includegraphics[width=0.95\linewidth]{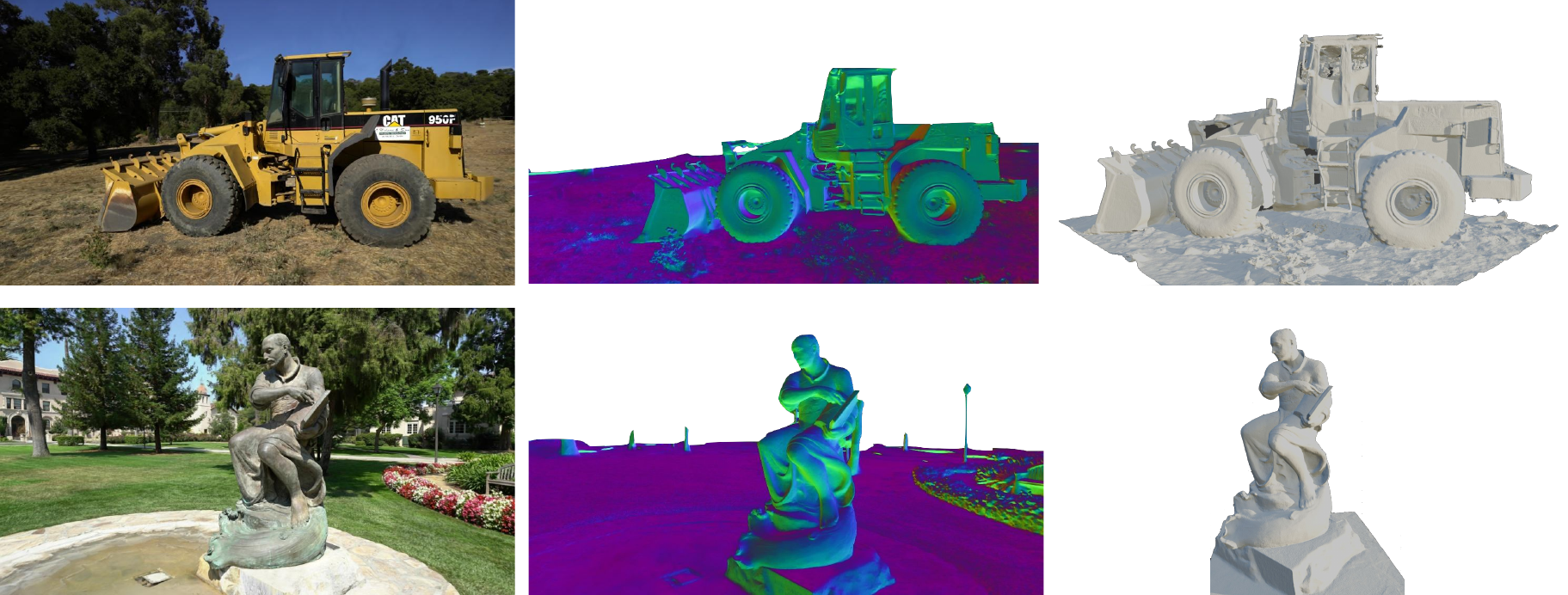}
    \caption{
        \textbf{Qualitative results on the Tanks and Temples dataset} Each row shows the input rendering, the detailed normal map, and the final high-quality mesh for the \textit{Caterpillar} and \textit{Ignatius} scenes.
    }
    \label{fig:tnt_qualitative}
    \vspace{-5mm}
\end{figure*}

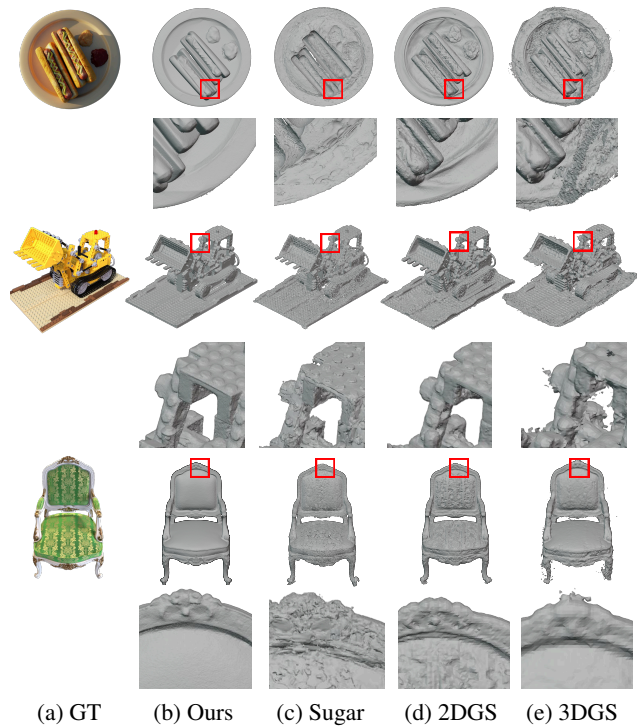
\begin{figure}[h]
    \centering
    \captionsetup{font=footnotesize}
    \vspace{7.5pt}
    \begin{overpic}[width=\linewidth]{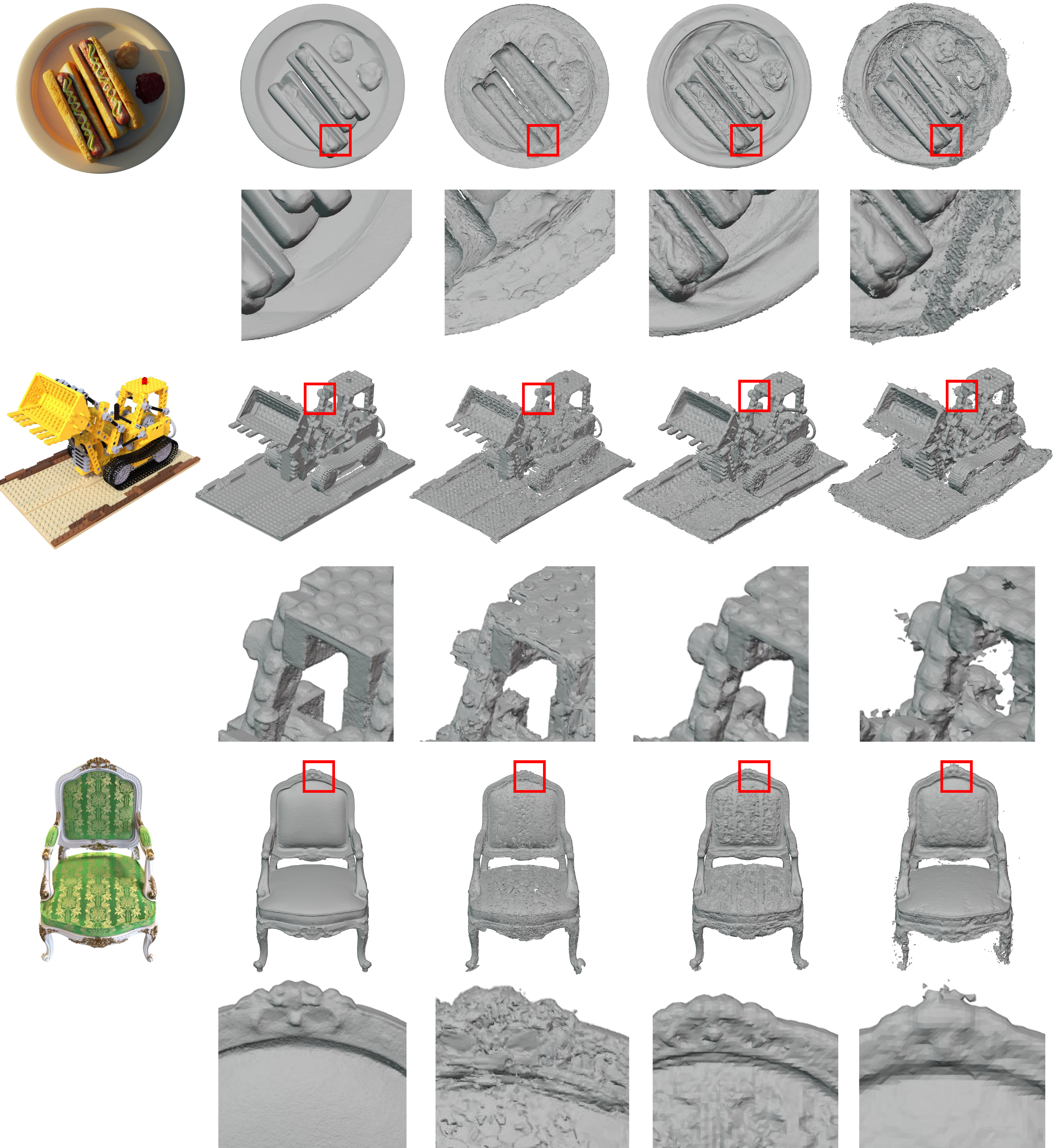}
        \put(4, -4.5){\small (a) GT}
        \put(21,-4.5){\small (b) Ours}
        \put(39, -4.5){\small (c) Sugar}
        \put(57.7, -4.5){\small (d) 2DGS}
        \put(75.5,-4.5){\small (e) 3DGS}
    \end{overpic}
    \vspace{-4pt}
    \caption{
        \textbf{Qualitative surface reconstruction on the Nerf-Synthetic dataset.} Red circles highlight areas where baseline methods like SuGaR (c) and 3DGS (e) suffer from prominent surface artifacts like bumpy textures. In contrast, our method (b) reconstructs a clean and smooth surface.
    }
    \label{fig:nerf}
    \vspace{-5mm}
\end{figure}

Our method demonstrates state-of-the-art performance in geometric reconstruction across multiple standard benchmarks. Table \ref{tab:dtu_geometry_comparison} shows that our method achieves a mean CD of 0.50 mm, outperforming all other explicit Gaussian-based reconstruction methods. This marks a significant 5.7\% improvement over the previous SOTA method, PGSR (0.53 mm), and even surpasses the computationally expensive implicit method, Neuralangelo (0.61 mm), by 18\%, all while requiring only 24 minutes of training. Qualitative comparisons in Figure \ref{fig:dtu4_1} visually confirms these quantitative strengths; our reconstructions are smoother and more complete, free from the "bumpy" artifacts common in baselines. For instance, in scan106 (the two birds), TriaGS improves accuracy over PGSR by 8.2\%. On the more challenging large-scale scenes of the Tanks and Temples dataset (Table \ref{tab:tnt}), our method achieves a mean F1-score of 0.49, which is highly competitive with PGSR (0.52). Figure~\ref{fig:tnt_qualitative} provides qualitative results, showcasing our ability to reconstruct detailed geometry on complex scenes. Notably, TriaGS achieves this result with a training time of only 0.8 hours, which is 33\% faster than PGSR's 1.2 hours. Furthermore, on the NeRF-Synthetic dataset, which features intricate shapes and fine details, Table \ref{tab:nerf_synthetic_quantiative} shows that our method achieves the best average Chamfer Distance of 0.76. We achieve an $8.4\%$ improvement over the best method, PGSR, which validates the high quality of our reconstructed geometry and its ability to capture complex topologies on synthetic data (as our qualitative result on Figure \ref{fig:nerf}).

\begin{table}[h]
\centering
\caption{\textbf{Quantitative results of F1 Score$\uparrow$ for reconstruction on Tanks and Temples dataset.}}
\resizebox{\columnwidth}{!}{%
\tabcolsep=3.5pt
\begin{tabular}{@{}l|cccccc|c|c@{}}
\textbf{Method} & \textbf{Barn} & \textbf{Caterpillar} & \textbf{Courthouse} & \textbf{Ignatius} & \textbf{Meeting Room} & \textbf{Truck} & \textbf{Mean} & \textbf{Time} \\
\hline
SuGaR & 0.14 & 0.16 & 0.08 & 0.33 & 0.15 & 0.26 & 0.19 & 2h \\
2D GS & 0.45 & 0.24 & 0.13 & 0.50 & 0.18 & 0.43 & 0.32 & \colorbox{colorSnd}{0.57h} \\
GOF & \colorbox{colorTrd}{0.51} & \colorbox{colorSnd}{0.41} & \colorbox{colorFst}{0.28} & \colorbox{colorTrd}{0.69} & \colorbox{colorSnd}{0.28} & \colorbox{colorTrd}{0.58} & \colorbox{colorTrd}{0.46} & 2h \\
RaDe-GS & 0.43 & 0.32 & \colorbox{colorSnd}{0.21} & \colorbox{colorTrd}{0.69} & \colorbox{colorTrd}{0.25} & 0.51 & 0.40 & \colorbox{colorFst}{11.5m} \\
PGSR & \colorbox{colorFst}{0.66} & \colorbox{colorFst}{0.44} & \colorbox{colorTrd}{0.20} & \colorbox{colorFst}{0.81} & \colorbox{colorFst}{0.33} & \colorbox{colorSnd}{0.66} & \colorbox{colorFst}{0.52} & 1.2h \\
\hline
Ours & \colorbox{colorSnd}{0.62} & \colorbox{colorTrd}{0.40} & \colorbox{colorTrd}{0.20} & \colorbox{colorSnd}{0.75} & \colorbox{colorSnd}{0.28} & \colorbox{colorFst}{0.71} & \colorbox{colorSnd}{0.49} & \colorbox{colorTrd}{0.8h} \\
\end{tabular}%
}
\label{tab:tnt}
\end{table}

\begin{table*}[h]
  \centering
  \caption{Quantitative comparison of Chamfer distance ($\times\ 10^{-2}$) comparison on NeRF-Synthetic dataset.}
  \label{tab:nerf_synthetic_quantiative}
  \resizebox{0.55\textwidth}{!}{
    \tabcolsep=3.5pt
    \begin{tabular}{@{}l|cccccccc|c@{}}
    Method & \sc{Chair} & \sc{Drums} & \sc{Ficus} & \sc{Hotdog} & \sc{Lego} & \sc{Materials} & \sc{Mic} & \sc{Ship} & AVG \\
    \midrule
    3DGS & 2.76 & 3.62 & 8.38 & 5.38 & 2.44 & 2.92 & 3.17 & 2.18 & 3.86 \\
    SuGaR & \colorbox{colorTrd}{0.74} & 1.01 & 0.67 & \colorbox{colorTrd}{1.47} & \colorbox{colorSnd}{0.70} & \colorbox{colorSnd}{0.75} & 0.79 & 1.25 & \colorbox{colorTrd}{0.92} \\
    2DGS & 0.95 & 1.57 & 2.80 & 2.10 & 1.04 & 1.09 & 1.33 & 1.92 & 1.60 \\
    GOF & 1.43 & \colorbox{colorTrd}{0.98} & \colorbox{colorTrd}{0.49} & 1.50 & \colorbox{colorFst}{0.61} & \colorbox{colorFst}{0.65} & \colorbox{colorFst}{0.65} & \colorbox{colorTrd}{1.21} & 0.94 \\
    PGSR & \colorbox{colorFst}{0.35} & \colorbox{colorSnd}{0.95} & \colorbox{colorFst}{0.40} & \colorbox{colorSnd}{1.20} & \colorbox{colorTrd}{0.73} & 1.56 & \colorbox{colorTrd}{0.73} & \colorbox{colorFst}{0.71} & \colorbox{colorSnd}{0.83} \\
    Ours & \colorbox{colorSnd}{0.50} & \colorbox{colorFst}{0.75} & \colorbox{colorSnd}{0.43} & \colorbox{colorFst}{1.12} & 0.74 & \colorbox{colorTrd}{1.09} & \colorbox{colorSnd}{0.69} & \colorbox{colorSnd}{0.77} & \colorbox{colorFst}{0.76} \\
    \bottomrule
    \end{tabular}
  }
\end{table*}

\begin{table}[h]
\centering
\caption{\textbf{Quantitative results on Mip-NeRF 360 \cite{mipnerf360} dataset}. We achieved competitive results, especially in the indoor scenes.}
\label{tab:mipnerf360_comparison}
\resizebox{\columnwidth}{!}{%
\begin{tabular}{l|ccc|ccc}
\toprule
& \multicolumn{3}{c|}{\textbf{Outdoor Scene}} & \multicolumn{3}{c}{\textbf{Indoor scene}} \\
\textbf{Method} & PSNR $\uparrow$ & SSIM $\uparrow$ & LPIPS $\downarrow$ & PSNR $\uparrow$ & SSIM $\uparrow$ & LPIPS $\downarrow$ \\
\midrule
NeRF & 21.46 & 0.458 & 0.515 & 26.84 & 0.79 & 0.37 \\
Deep Blending & 21.54 & 0.524 & 0.364 & 26.4 & 0.844 & 0.261 \\
Instant NGP & 22.9 & 0.566 & 0.371 & 29.15 & 0.88 & 0.216 \\
MERF & 23.19 & 0.616 & 0.343 & 27.8 & 0.855 & 0.271 \\
\hline
3D GS & 24.64 & 0.731 & 0.234 & 30.41 & 0.92 & 0.189 \\
Mip-Splatting & 24.65 & 0.729 & 0.245 & \colorbox{colorFst}{30.9} & 0.921 & 0.194 \\
2D GS & 24.34 & 0.717 & 0.246 & 30.4 & 0.916 & 0.195 \\
GOF & \colorbox{colorTrd}{24.82} & 0.75 & \colorbox{colorSnd}{0.202} & \colorbox{colorTrd}{30.79} & 0.924 & 0.184 \\
RaDe-GS & \colorbox{colorFst}{25.17} & \colorbox{colorFst}{0.764} & \colorbox{colorFst}{0.199} & 30.74 & \colorbox{colorTrd}{0.928} & \colorbox{colorTrd}{0.165} \\
PGSR & 24.76 & \colorbox{colorSnd}{0.752} & \colorbox{colorTrd}{0.203} & 30.36 & \colorbox{colorFst}{0.934} & \colorbox{colorFst}{0.147} \\
\textbf{Ours} & \colorbox{colorSnd}{24.95} & \colorbox{colorTrd}{0.751} & 0.210 & \colorbox{colorSnd}{30.89} & \colorbox{colorSnd}{0.930} & \colorbox{colorSnd}{0.164} \\
\bottomrule
\end{tabular}%
}
\end{table}
\begin{figure}[h]
    \centering
    \captionsetup{font=footnotesize}
    \includegraphics[width=\linewidth]{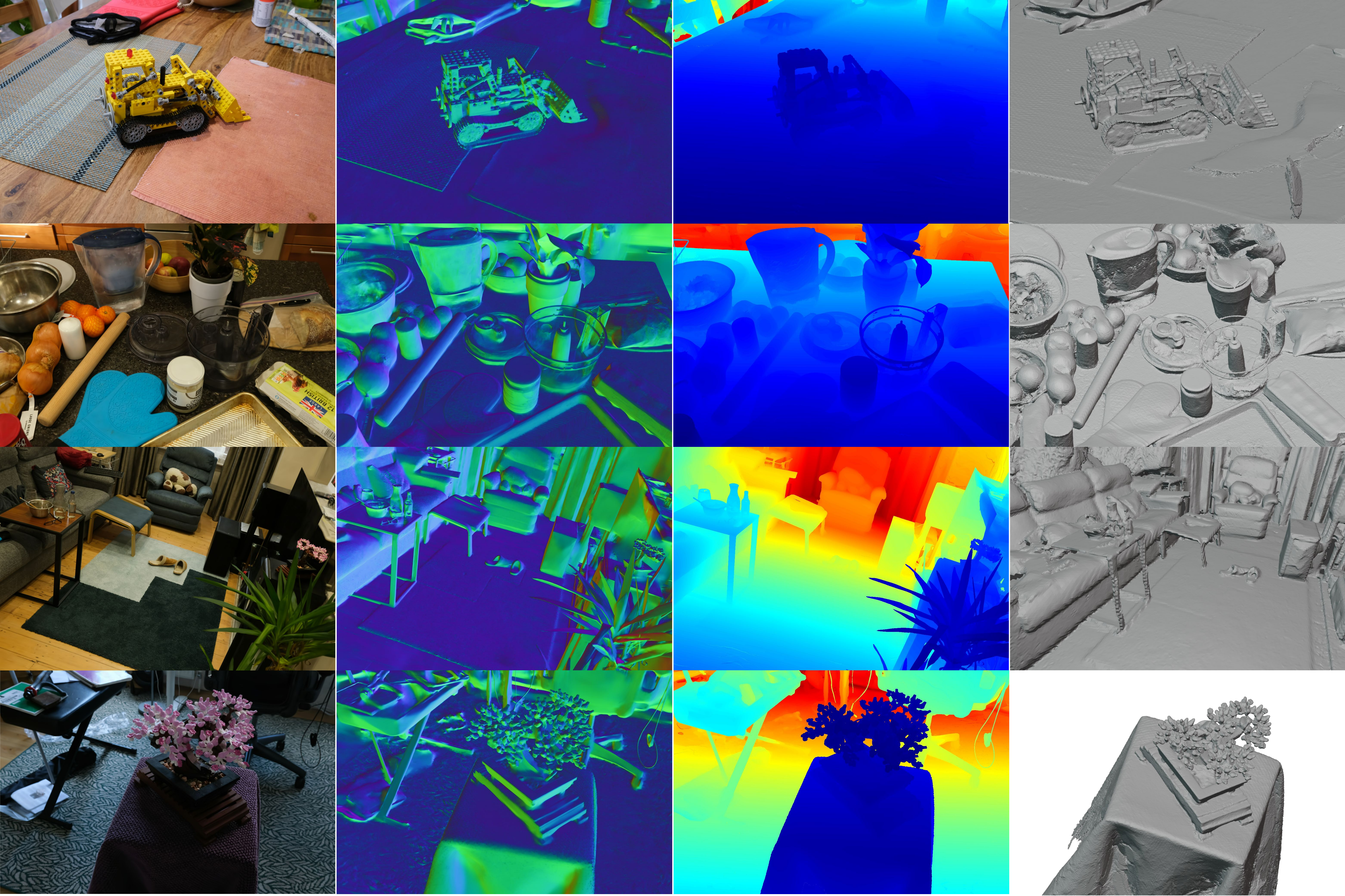}
    \caption{
        \textbf{High-fidelity reconstruction results on the Mip-NeRF 360 dataset \cite{mipnerf360}} From left to right: rendered view, normal map, depth map, and the final extracted mesh. Our approach consistently produces high-quality geometry for complex scenes, including \textit{kitchen}, \textit{Counter}, \textit{Room}, and \textit{Bonsai} (top to bottom).
    }
    \label{fig:mip360_qualitative}
\end{figure}

\subsection{Novel View Synthesis Quality}
We assess the quality of novel view synthesis using the Mip-NeRF 360 dataset \cite{mipnerf360} to ensure that our geometric regularization does not compromise rendering quality. As shown in Table \ref{tab:mipnerf360_comparison}, our method performs on par with the original 3DGS and others across PSNR, SSIM, and LPIPS metrics. As shown in Figure \ref{fig:mip360_qualitative}, TriaGS confirms that our global geometric loss successfully improves surface quality without impairing the rendering capabilities.

\subsection{Ablation Studies}

\begin{table}[h]
  \centering
  \caption{Ablation study on the number of neighboring views ($k$). The $k=1$ setting corresponds to pairwise consistency. Increasing k improves the F1-Score, with $k=12$ achieving the best balance.}
  \label{tab:ablation_k_views}
  \resizebox{\columnwidth}{!}{%
    \begin{tabular}{l|cc|c}
    \toprule
    \textbf{Variant} & \textbf{PSNR $\uparrow$} & \textbf{F1-Score (5mm) $\uparrow$} & \textbf{Iterations/s $\uparrow$} \\
    \midrule
    $k=1$ (Pairwise) & 23.21 & 0.57 & 12.8 \\
    $k=4$ & \colorbox{colorSnd}{24.11} & 0.62 &  11.6 \\
    $k=8$ & \colorbox{colorFst}{24.38} & \colorbox{colorSnd}{0.66} & 10.6 \\
    $k=12$ & 23.94 & \colorbox{colorFst}{0.71} & 9.8 \\
    \bottomrule
    \end{tabular}%
    }
\end{table}

\begin{figure}[h]
    \centering
    \includegraphics[width=0.95\linewidth]{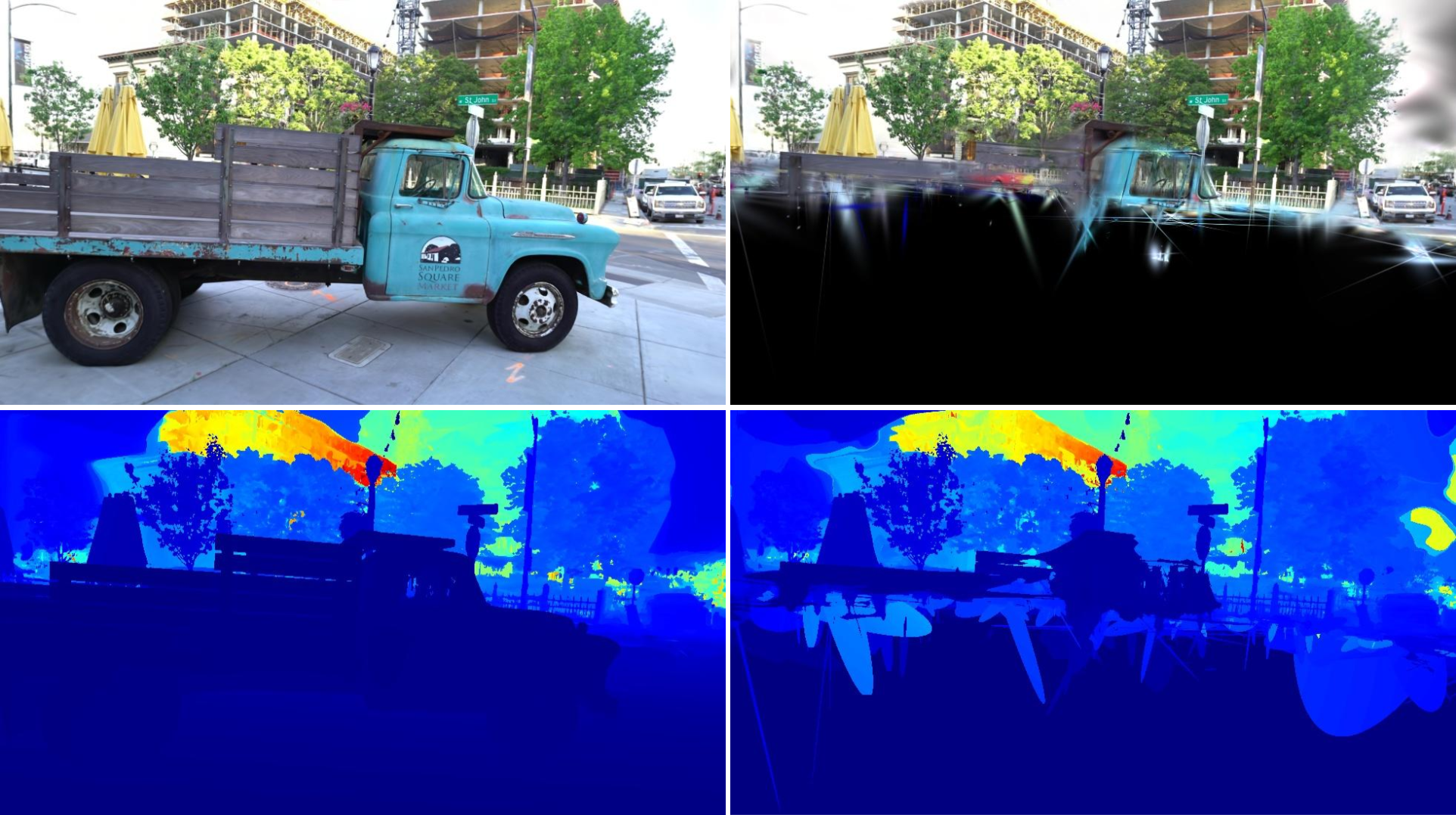}
    \captionsetup{font=footnotesize}
    \caption{
        \textbf{Ablation on Loss Function}. Comparison of models trained with Geman-McClure (left column) vs. $L_2$ loss (right column). \del{ The $L_2$ loss leads to optimization divergence, resulting in a corrupted depth map and a failed render, while our proposed loss remains stable.}
    }
    \label{fig:ablation_loss_function}
\end{figure}

We conduct ablation studies on the Truck scene from the Tanks and Temples dataset to validate our design choices. Our hypothesis posits that multi-view consensus is superior to local, pairwise checks. To test this, we evaluate our TGGC loss with a varying number of neighboring views, $k$. As shown in Table \ref{tab:ablation_k_views}, performance improves significantly as more views are incorporated. We select $k = 12$ as our final configuration, as it achieves the best balance of F1-Score, and further increasing $k$ resulted in higher computational costs. In addition, we employ the Geman-McClure loss to handle noisy geometric estimates present in the early stages of optimization. Figure \ref{fig:ablation_loss_function} illustrates a direct comparison: a model trained with a standard $L_2$ loss suffers from catastrophic optimization divergence due to its sensitivity to outliers, leading to gradient explosion and complete reconstruction failure. This demonstrates that a robust loss function is not merely an enhancement but a crucial component for the success of our triangulation-guided method.

\vspace{-15pt}
\section{Conclusions}
In our work, we have presented Triangulation-Guided Geometric Consistency, a self-supervised regularizer that enforces a global geometric consensus in 3DGS via a differentiable multi-view triangulation module. Our method achieves high-fidelity surface reconstruction by utilizing only the scene's intrinsic geometric constraints. However, as a geometry-focused method, our performance is sensitive to input data quality. The mesh reconstruction can degrade with camera pose inaccuracies or in sparsely viewed scenes. These limitations highlight promising directions for future research. The triangulation residual itself could serve as a signal for pose refinement, turning sensitivity into potential strength. Furthermore, integrating our constraint with hybrid implicit-explicit models could provide robust priors where multi-view evidence is weak, advancing the state-of-the-art in reconstruction accuracy and completeness.

\newpage

{\small
\bibliographystyle{ieee_fullname}
\bibliography{egbib}
}

\end{document}